\documentclass[conference]{IEEEtran}
\IEEEoverridecommandlockouts
\usepackage[round, sort&compress]{natbib}
\usepackage{amsmath,amssymb,amsfonts}
\usepackage{algorithmic}
\usepackage{graphicx}
\usepackage{textcomp}
\usepackage{xcolor}
\usepackage{url}
\usepackage{hyperref}
\hypersetup{
  colorlinks   = true, 
  urlcolor     = blue, 
  linkcolor    = black, 
  citecolor    = black 
}
\usepackage{booktabs, multirow}
\usepackage{makecell}

\def\BibTeX{{\rm B\kern-.05em{\sc i\kern-.025em b}\kern-.08em
    T\kern-.1667em\lower.7ex\hbox{E}\kern-.125emX}}
\begin{document}

\title{Controlled Text Generation using T5 based Encoder-Decoder Soft Prompt Tuning and Analysis of the Utility of Generated Text in AI. \\
\thanks{ \textsuperscript{*}Note: The code-base for the models developed can be found in \url{https://github.com/damith92/T5_encoder_decoder_prompt_tuning_for_text_generation}
}
}

\author{\IEEEauthorblockN{Damith Chamalke Senadeera}
\IEEEauthorblockA{
\textit{Queen Mary University of London}\\
d.c.senadeera@se21.qmul.ac.uk}
\and
\IEEEauthorblockN{Dr. Julia Ive}
\IEEEauthorblockA{
\textit{Queen Mary University of London}\\
j.ive@qmul.ac.uk}
}

\maketitle

\thispagestyle{plain}
\pagestyle{plain}

\begin{abstract}

Controlled text generation is a very important task in the arena of natural language processing due to its promising applications. In order to achieve this task we mainly introduce the novel soft prompt tuning method of using soft prompts at both encoder and decoder levels together in a T5 model and investigate the performance as the behaviour of an additional soft prompt related to the decoder of a T5 model in controlled text generation remained unexplored. Then we also investigate the feasibility of steering the output of this extended soft prompted T5 model at decoder level and finally analyse the utility of generated text to be used in AI related tasks such as training AI models with an interpretability analysis of the classifier trained with synthetic text, as there is a lack of proper analysis of methodologies in generating properly labelled data to be utilized in AI tasks. 

Through the performed in-depth intrinsic and extrinsic evaluations of this generation model along with the artificially generated data, we found that this model produced better results compared to the T5 model with a single soft prompt at encoder level and the sentiment classifier trained using this artificially generated data can produce comparable classification results to the results of a classifier trained with real labelled data and also the classifier decision is interpretable with respect to the input text content.

\end{abstract}


\section{Introduction}

As one of the main two branches in Natural Language Processing (NLP), Natural Language Generation (NLG), aims on producing human understandable natural language text from internal machine representations \citep{ horacek2001building}. In this field of NLG, controlled text generation meeting some pre-specified attributes, aims to generate text which adheres to a given attribute like an emotion or a sentiment and it has been a demanding and a challenging task for machines. Due to this complexity in this problem and also due to its promising applications in the field of natural language processing with applications to tasks like review generation where you need controlled text preserving the context according to the requirements, this problem has caught the attention of the NLP research community in the recent past \citep{ zhang2022survey}. 

With the rapid recent advancements in the field of Deep Learning, NLP related Deep Learning research has produced encouraging breakthroughs in the field including in NLG. But a major bottleneck faced when it comes to training Deep Learning models in NLP is the lack of large amounts of labelled data \citep{bhattacharyya2012natural}. As one of the solutions to this problem of scarcity of labelled data, recent studies have risen to explore the feasibility of employing text generation mechanisms to produce artificially generated text data which adheres to a specific label or an attribute to be used in training machine leaning / deep learning models \citep{ive2020generation}.

Numerous studies on controllable text generation using Deep Learning models also have taken place in the recent years and previous work done by \citep{zhang2020pointer, wang2021mention, liu2021dexperts} make use of pre-trained transformer-based \citep{vaswani2017attention} large language models such as BERT related models \citep{devlin2018bert} and GPT-2 \citep{radford2019language} and the generation is steered according to pre-specified objectives. In almost all of these studies the researchers have made use of the most widely used approach of fine-tuning all the parameters of the entire large pre-trained model which requires a considerable amount of time and computational resources as these pre-trained large language models consist of 100s of millions if not billions of parameters \citep{ yang2021fudge}. 

As a substitute for fine tuning all the parameters in large pre-trained language models, prompt-based learning techniques have emerged as a popular paradigm in the more recent research work. According to \citep{lester2021power}, “Prompting is the approach of adding extra information for the model to condition on during its generation of Y” where Y is the output and generally it is achieved by appending or prepending a series of text tokens to the input text in order to reformulate whatever the original task (be it classification or generation) into a masked language modelling problem. This complete process of reshaping the input text with a prefix or a suffix text template is known as hard prompting.

\begin{figure}[htbp]
\centerline{\includegraphics[height=0.5\linewidth]{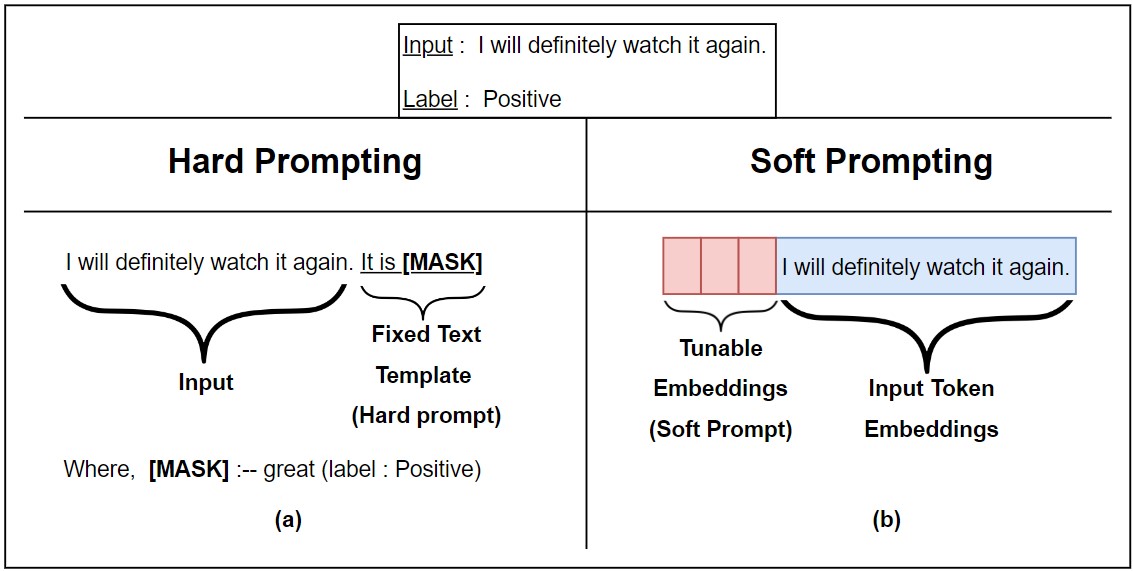}}
\caption{The comparison between the hard prompting (part (a)) where the input is transformed into a predefined text template according to the attributes of the input (label) and soft prompting (part (b)) where a tunable piece of embedding is prepended to the input}
\label{fig0}
\end{figure}

Compared to engineering these hard prompts as seen in figure \ref{fig0} part (a), another technique of prompt tuning using soft prompts where attaching a piece of learnable input embeddings to the main input to the model as shown in figure \ref{fig0} part (b) and optimizing those embeddings has shown to achieve results better than hard prompting and on-par results with complete model fine tuning for NLP classification tasks such as paraphrase detection and generation tasks such as question answering \citep{lester2021power}. But, for generation tasks, although the decoder of an encoder-decoder architecture plays an important role, the use of a soft prompt at the decoder level of this encoder-decoder model (specifically T5) has not been explored. 

Therefore this project intends to introduce a novel method of soft prompt tuning incorporating a soft prompt at decoder level of an encoder-decoder based deep learning architecture (for specifically T5 model \citep{ raffel2020exploring}) along with an encoder soft prompt for a controlled text generation task as decoder plays an important role in text generation attending the input query to encoder embeddings and learning how to generate the next token attending to the previous tokens in a transformer based encoder-decoder deep learning architecture \citep{vaswani2017attention} and compare the results to that of obtained from the baseline models of T5 with a single soft prompt at the encoder level \citep{ lester2021power} and soft prompted GPT-2 at input level \citep{ yang2022tailor}. Also we explore the usage of steering method discussed in \citep{ liu2021dexperts} to steer the text generation at decoder level using the modified T5 model with encoder and decoder level soft prompts as in the work of \citep{ liu2021dexperts}, the researchers have only investigated the steering of a decoder based model (GPT-2) that too only with static hard prompts. The experiments on these developed methodologies are carried out using a filtered set of Amazon product review data \citep{ni2019justifying} on Movies and TV related products. Finally classification performance of real labelled data are analysed on a sentiment classifier trained with artificial product reviews obtained from the best models developed.

\begin{figure}[htbp]
\centerline{\includegraphics[height=0.7\linewidth]{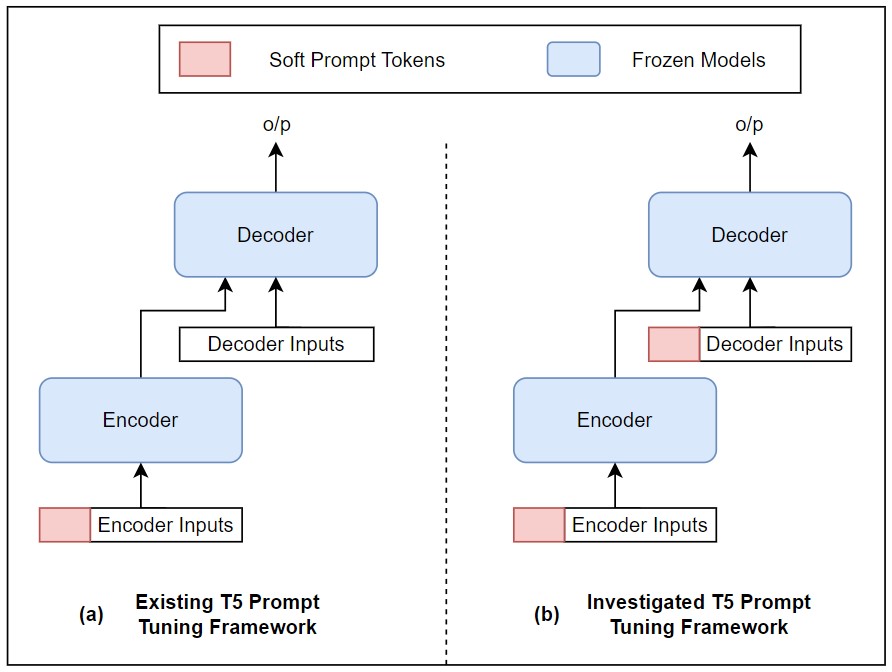}}
\caption{The comparison between the previous T5 prompt tuning method (part (a)) and the introduced T5 prompt tuning method (part (b))}
\label{fig1}
\end{figure}

The main goals of this project are to (1) introduce and implement a novel method for an extended soft prompt tuning architecture for T5 model incorporating soft prompts at both encoder and decoder levels and investigate the performance compared with the existing baseline soft prompt architectures of \citep{ lester2021power} and \citep{ yang2022tailor} and analyse the results; (2) investigate the feasibility of steering the output of an encoder-decoder architecture at decoder level using expert \& anti-expert methodology in steering  introduced by \citep{ liu2021dexperts} ; and (3) explore the utility of artificially generated text with an in depth intrinsic and extrinsic analysis to be used in AI related tasks such as training AI models with artificial data and testing their performance with respect to real sentiment labelled data with a classifier interpretability analysis as well.

As the main contributions, we were able to introduce a novel method of soft prompt tuning incorporating an additional soft prompt at decoder level of T5, which produced superior performance compared to the existing baseline model of T5 with a single encoder soft prompt which was mainly evident through the extrinsic evaluation and also we were able to show that the best classifier trained with artificial data produced from the proposed novel model, produce not just random classification results but interpretable results based on the different positive and negative words of the input text.

\section{Related Work}

\subsection{Attribute Based Controlled Text Generation via Full Model Fine-Tuning}

The goal of attribute-based controlled text generation is to produce natural language sentences that meet criteria like themes and emotions. One of the main requirements of real-world writings is to have a carefully preserved context in the text and to achieve that objective, machine-based text generation systems in theory should be able to produce controlled text by carefully manipulating the input features to the system \citep{zhang2022survey}. But in reality, this precise manipulation of input features to the text generation system is quite a complex and a challenging task as there is no clear-cut mechanism defined on how the manipulation should occur \citep{zhang2022survey}. 

In the work of \citep{zhang2020pointer}, they have introduced a progressive insertion-based transformer approach for hard constrained text generation by progressively inserting new tokens between existing tokens in a parallel manner arguing that large language models cannot be utilized off the shelf for controlled text generation although they can be used for open ended text generation. One large bottleneck of this work is that to train and generate text using this system, the users must prepare the lexical restrictions manually. 

Simplifying this bottleneck, the authors in \citep{ive2020generation} have developed an attribute based controlled text generation system which uses RAKE algorithm to extract key words from the training data and then using these keywords a transformer model has been trained to generate text data (specifically electronic health records). In the end the artificially generated data has been test on the preservation of various attributes they have considered during the generation of those artificial electronic health records. Again, for all the work discussed above including this method, needs the full large language model to be fine-tuned completely. 

\subsection{Decoder Level Steering Based Controlled Text Generation}

To avoid the complete fine tuning of the large language models, studies in steering the generation of text at decoding time with respect to some pre-defined attributes have emerged recently. In the work of \citep{dathathri2019plug}, first the authors have trained a smaller attribute classifier compared to a large language model and then the text generation from a large language model is guided with the help of this smaller attribute classifier at the decoding stage without further modifications to the pre-trained large language model. As a limitation of this method, \citep{yang2022tailor} have reported a considerable decrease in the fluency of generation (a considerably higher perplexity) and an increased inference time compared to the other existing latest methods.

Getting insights from the previous approach, another novel steering approach has been developed in the work of \citep{ liu2021dexperts}, where the researchers use two pre-trained large language models designated as an “expert” which guides the output towards the favorable behavior and the other designated as the “anti-expert” which guides the output to be away from the un-favourable behavior during the generation step at the decoder of the base language model. Under this aggregated steering approach, authors in \citep{ liu2021dexperts} argue that the generated tokens will get a higher probability only if they are considered highly likely by the expert and not very likely by the anti-expert.

If we dive deep into the working principal of the steering methodology introduced by \citep{ liu2021dexperts}, during the decoding stage where the output is generated by the base language model $M$ by combining the output of the expert model $M^{+}$ and the anti-expert model $M^{-}$, at each time step $t$ of the decoding stage the language models $M$, $M^{+}$ and $M^{-}$ are conditioned on the hard prompt $x_{<t}$ to obtain $z_{t}$, $z_{t}^{+}$ and $z_{t}^{-}$. These produced probabilities are then combined to get the final probability of generation as shown below .
\begin{equation}
 \widetilde{P}(X_{t}|x_{<t}) = softmax(z_{t}+\alpha (z_{t}^{+} - z_{t}^{-})) \label{eq1}
\end{equation}
Where $\alpha$ is the tunable hyperparameter in equation \eqref{eq1} which can be used to modify the steering. But again, as a bottleneck here for this approach we need to have 2 fully fine-tuned large language models according to the steering objective as the expert and the anti-expert to guide the generation.

\subsection{Prompt Tuning Based Controlled Text Generation}

Prompt tuning is one of the latest paradigms in NLP to assist a large pre-trained language model to attend to different downstream tasks via attaching a specific prefix or a suffix with the input to the model according to the downstream task \citep{ zhang2022survey}. As visualized in figure \ref{fig0} there are 2 main methods of prompting, namely hard prompting and soft prompting where hard prompting is achieved by transforming the input text with the help of a prefix or a suffix template with a natural language task description (as seen in part (a) of the figure \ref{fig0}) and where soft prompting is archived by prepending a task-specific tunable tokens to the input (as seen in part (b) of the figure \ref{fig0}) \citep{lester2021power}.

In previous studies, methods in prompt engineering where prompts were discrete word templates (hard prompts) have been investigated by \citep{shin2020autoprompt, liu2021gpt} as these hard prompts are interpretable. Due to this advantage, human understandable prompt formulation was possible. But due to the inherent complexity in finding the best hard prompt for a particular task given the large space where hard prompts can be formulated, the research has been driven towards developing a more flexible soft prompt paradigm where the prompt is formulated as a set of continuous vectors which can be prepended to the input of the model \citep{lester2021power}. In contrast to the hard prompts which may have trouble being optimized, soft prompts may be trained directly using backpropagation for a specific downstream task. 

In \citep{lester2021power} the authors explore the usage of a soft prompt embedded with the input to the T5 model only at encoder level \citep{raffel2020exploring} which is a transformer encoder-decoder \citep{ vaswani2017attention} based large language model, in order to condition the frozen T5 model for various down-stream task by only just fine tuning the embeddings related to soft prompt through backpropagation while keeping the large language model frozen. The authors have shown that this method can achieve on-par results to full parameter tuning of a model for tasks such as question answering and paraphrase detection by just fine-tuning an amount of parameters less than 0.01\% of the total large model parameters  \citep{lester2021power}.

The authors in \citep{yang2022tailor} have made use of the same principal of fine tuning a soft prompt prepending it to the input of the GPT-2 model \citep{radford2019language} which is only a decoder-based transformer model to generate text satisfying various attributes. They also have experimented on how to combine different soft prompts to produce text satisfying multiple attributes using the GPT-2 model as well. But again, this study lacks comparison of the usage of the soft prompts in an encoder-decoder based architecture compared only to a decoder-based architecture for a text generation task. Also, our experiments reported in table \ref{table3} show that in the task of learning generators based on sentiments related to Movie and TV product reviews, T5 based single input soft prompt tuning discussed by \citep{lester2021power} outperforms the GPT-2 based generations.

\section{Preliminaries}

\subsection{T5 Base Model}

According to the authors in \citep{raffel2020exploring} the initial T5 model which is an encoder-decoder based model, has been originally trained on an objective of “span corruption” which means the model is pre-trained based on the task of “reconstructing” the spans which are masked in the input tokens where the masking is done with the help of special sentinel tokens. More specifically the input training text consists of some sentinel tokens and the output target text will contain the tokens which were masked separated by their corresponding sentinel tokens and a final sentinel token to mark the end of the stream. For example if the original sentence is “I went home after a tedious long day”, a possible training input will be masked with sentinel tokens to look like “I went $<X> $ after a $<Y>$  day” and the target output would have been “$<X>$ home $<Y>$ tedious long $<Z>$” where the sentinel token $<Z>$ marks the end of the stream.

But in \citep{lester2021power}, the researchers argued that this learning objective is not a very good learning objective to pre-train a frozen model which is to be conditioned through soft prompt tuning as the model trained in this manner has not been exposed to real natural text inputs without sentinel tokens. As an alternative, the authors in \citep{lester2021power} has obtained the original T5 model and trained that for an additional 100,000 steps with a Language Modelling (LM) objective of predicting the next token and they showed that this newly adapted T5 model worked better for prompt tuning tasks compared to the original T5 model.

Therefore, for all our experiments we have used this “LM Adapted” T5 model which has been released publicly by the authors of \citep{lester2021power} which can be found in \citep{t5lmadapthf}. We used the small version of this model with only 60 million parameters (compared to the largest model with 11 billion parameters) due to the resource constraints faced.

\subsection{Data Set}

To experiment with, we chose the Amazon Product Review dataset \citep{ni2019justifying} which is a publicly available dataset of 233.1 million text reviews for products from Amazon.com including metadata such as the star ratings. Due to the public availability and also due to the ability to categorize the reviews into positive and negative sentiment based on the star ratings given, many machine learning and deep learning researchers including \citep{zhang2015character, haque2018sentiment, jagdale2019sentiment} have used data extracted from this dataset, labelled into different sentiments according to the star ratings, in order to train and test sentiment classification models. 

Therefore, to conduct our experiments we use data extracted from the Movie and TV product related reviews subset of this Amazon Product Review dataset which contains 8.7 million reviews and those reviews are labelled positive if the star rating is more than 3 and negative if it’s less than 3 following the same methodology \citep{zhang2015character} used to prepare labelled data out of this Amazon Product review dataset. More details on how data were selected to train different models and filtering steps can be found in the data pre-processing section.

\section{Methodology}

\subsection{Data Pre-Processing}

In order to experiment on the proposed models, Amazon Product Reviews related to Movie and TV products are used and pre-processed using the Python libraries of NumPy and Pandas \citep{pypandaspkg, pynumpypkg}. First, since the data seem to contain a lot of html tags through visual inspection as this dataset is scraped from the web, it was decided to parse the data through an xml parser and filter all the xml/html tags and related words using the Beautiful Soup python package \citep{pybeautifulspkg}. Then, to facilitate this objective further, all the non-word entities like html entities, un-natural non-dictionary words were filtered using regular expressions. Afterwards, all the contractions in English like “can’t”, “I’m”, etc which were detected through the python package “contractions” \citep{pycontractionpkg} were also removed, to try to maintain the uniformity in the language. After that, all the punctuations and special characters were also removed from the data as it was observed in many reviews there were repeated sequences of punctuations such as “.......”, “*******”, etc to again maintain the focus towards the LM objective.

Following the method used by \citep{zhang2015character}, these text reviews were then labelled into positive and negative class labels where, if the star rating was more than 3 the reviews were labelled positive and if the star rating was less than 3 the reviews were labelled negative. All the reviews with no star ratings and with a star rating of 3 were discarded. After all these filtrations, the dataset got reduced to 7.05 million records where there were 6.1 million positive reviews and 914K negative reviews. Afterwards, to avoid data leakage between different models and also within the same model in training, validation, and testing sets, all the exact duplicates among these reviews were searched and removed too. After the removal of the duplicates, there were 5.5 million positive reviews while there were only 846K negative reviews.

\begin{table}[htbp]
\caption{The common structure of the datasets}
\begin{center}
\begin{tabular}{| p{2.5cm} | l| l| l|}
\hline
&\textbf{Training Set} & \textbf{Validation Set} & \textbf{Testing Set} \\
\hline
Positive Sentiment Labelled Data & 25K & 5K & 5K \\
Negative Sentiment Labelled Data & 25K & 5K & 5K \\
\hline
Total & 50K & 10K & 10K \\
\hline
\end{tabular}
\label{table1}
\end{center}
\end{table}

Since this cleaned dataset has a high class imbalance and the amount of data is also too large to conduct experiments within the available time-frame, following the method discussed by \citep{keung2020multilingual} to downsample the data in order to obtain a class balanced data set, the token lengths of each reviews were calculated and each review was given a weight according to the distribution of the token lengths in the positive review set and the negative review set separately with the objective of preserving the original review length distribution in the downsampled data. Preserving this distribution of sentence lengths, down sampling was done without replacement to form 3 different datasets which are class balanced (which will be known as Dataset-1, Dataset-2 and Dataset-3 from here on-wards) with the common structure shown in table \ref{table1} for the experiments.

\subsection{Model Formulations }

\subsubsection{Sentiment Classification Model}

To be used as the sentiment classifier, DistilBERT model \citep{sanh2019distilbert} which is a transformer encoder based bidirectional model which conditions jointly on the context of right and left in all the layers of the transformer encoder which has been transfer learned from the bigger model BERT \citep{ devlin2018bert} through knowledge distillation was chosen due to it’s light weight and comparable performance in classification tasks. 

\subsubsection{Text Generation Models Designed with only Input Soft Prompt}

Deviating from fine-tuning the whole large language model, for the base T5 and the GPT-2 models, only the soft prompt parameters will be updated during training. As shown in Figure \ref{fig1}- part (a), a single soft prompt $S_{in}$ with length $l_{s}$ is initialized with sampled vocabulary from the input embedding layer to the model as \citep{lester2021power} has shown that it’s better to initialize the soft prompts this way rather than initializing randomly where $ S_{in} \in \mathbb{R}^{l_{s} \times d_{emb}} $ when the dimensions of the input word embeddings are represented by $ d_{emb} $. 

Then, for an input sentence labeled positive or negative $x = \{ x_{1}, ..., x_{n} \}$ with length $n$ , the word embeddings $ X_{emb} \in \mathbb{R}^{n \times d_{emb}} $ are obtained and joined together with the respective soft prompt to form the input embeddings to the model $ [S_{in} ; X_{emb}] \in \mathbb{R}^{(l_{s} + n) \times d_{emb}} $. Afterwards, the language modeling (LM) objective of predicting the next token given a sequence of input tokens is used to train the respective soft prompt keeping the large language model frozen. This learning objective is characterized by the equation \eqref{eq2}

\begin{equation}
L = \sum_{t=1}^{n}logP_{\theta_{m};\theta_{S_{in}}}(x_{t} | [S_{in} ; x_{1}, ..., x_{t-1}]) \label{eq2}
\end{equation}

where $\theta_{m}$ denotes the frozen parameters of the large language model (T5 or GPT-2) and $\theta_{S_{in}}$ denotes the input soft prompt parameters which are fine-tuned.

During the inference stage, $\theta_{S_{in}}$ is concatenated with the embeddings of the few input tokens introduced to initiate generation and generation of new logits is carried out based on this concatenated embeddings.

\subsubsection{T5 Model with Soft Prompts at the Encoder and Decoder}

As shown in Figure \ref{fig1}- part (b), two soft prompts, $S_{en}$ with length $l_{en}$ for the encoder and $S_{de}$ with length $l_{de}$ for the decoder are initialized with sampled vocabulary from the encoder embedding layer and the decoder embedding layer of the T5 model respectively where $ S_{en} \in \mathbb{R}^{l_{en} \times d_{en-emb}} $ given the dimensions of the encoder word embeddings are represented by $ d_{en-emb} $ and $ S_{de} \in \mathbb{R}^{l_{de} \times d_{de-emb}} $ given the dimensions of the decoder word embeddings are represented by $ d_{de-emb} $. 

Next, similar to the earlier approach, the encoder word embeddings for the input sentence $x = \{x_{1}, ..., x_{n}\}$  with length $n$ is obtained and concatenated with the respective encoder soft prompt and the decoder word embeddings are also obtained for the same input sentence and concatenated with the decoder soft prompt. Then, to train the encoder and the decoder soft prompts keeping the T5 model frozen, the language modeling (LM) objective discussed earlier is used where a modification is introduced to learn the decoder soft prompt additionally. The modified learning objective to include both the encoder and decoder soft prompts is elaborated in the equation \eqref{eq3}

\begin{equation}
\begin{split}
L = \sum_{t=1}^{n}logP_{\theta_{m};\theta_{S_{en}}; \theta_{S_{de}}}(x_{t} | [S_{en} ; x_{1}, ..., x_{t-1}], \\ 
[S_{de} ; x_{1}, ..., x_{t-1}])
\end{split}
\label{eq3}
\end{equation}

where $\theta_{m}$ denotes the frozen parameters of the T5 model and $\theta_{S_{en}}$ and $\theta_{S_{de}}$ denotes the encoder and decoder soft prompt parameters which are fine-tuned.

At the inference stage, $\theta_{S_{en}}$ is concatenated with the embeddings of the input tokens introduced for generation and the output hidden states of the encoder which are used in the decoder, is calculated based on this concatenated embeddings. Then, $\theta_{S_{de}}$ is concatenated with the decoder start-sequence token embeddings and attending to the output hidden states of the encoder, the generation of the new logits are carried out.

\subsubsection{Expert \& Anti Expert Steering for T5 model with Encoder-Decoder Soft Prompts}

Following the inspiration gained by the method discussed in \citep{liu2021dexperts} where the output of a Language Model (LM) is steered during decoder level using the logit probabilities obtained through an "expert" model steering the output towards the favourable behaviour and another "anti-expert" model which steers the output away from a designated un-favourable behaviour, as shown in Figure \ref{fig3} in Appendix \ref{ap0} section \ref{ap01}, the output at the decoder stage of the base T5 model which is same as the expert T5 model trained with encoder and decoder soft prompts was tried to steer away from the unfavourable behaviour of the anti-expert T5 model also trained with encoder and decoder soft prompts. Incorporating $\alpha$ as the tunable hyperparameter as seen in equation \eqref{eq1} steering could be modified accordingly.

For the positive review generation, as the base and the expert LM, the T5 model trained to generate positive reviews with encoder and decoder soft prompts was used while the T5 model trained to generate negative reviews with encoder and decoder soft prompts was used as the anti-expert. For the negative review generation, the base and the expert LM was swapped with the anti-expert model which was used for positive review generation.

\subsection{Experiments}

All the experiments related to model training, evaluation were conducted using the 1.13.0 version of the Python framework PyTorch \citep{pytorchpkg} and all the transformer related pre-trained large language models developed using PyTorch framework were obtained through the 4.24.0 version of the Hugging Face python library \citep{pyhfpkg} which were then modified according to the requirements.

The initial DistilBERT sentiment classifier model which was used to evaluate the sentiment classification results of the real labelled data was trained and evaluated using the Dateset-1. For all the generation models with soft prompts, a soft prompt token length of 20 was chosen considering the the fact that \citep{lester2021power}  had shown that, after a soft prompt token length of 20, the results of the earlier soft prompt model developed by them did not improve drastically. All these generation models were trained and evaluated using the Dateset-2 for the intrinsic evaluation where the DistilBERT model which was trained earlier using real labelled data was utilized for testing the classification accuracy of the generated data.

Finally, using the 3 models which incorporated soft prompts with T5 model, artificial product reviews were generated using the Dataset-3 and for the purpose of extrinsic evaluation, new DistilBERT models were trained using the artificially generated labelled data and were evaluated on the test set of the real labelled data of Dataset-1. More details regarding these experiments can be found in Appendix \ref{ap1} .

\section{Results and Discussions}

\subsection{Evaluation of Sentiment Classifier Model Trained with Real Data}

To evaluate the DistilBERT model finetuned with the real labelled data, accuracy of classification along with the per label F1 Scores were used. Since data is class balanced, the model testing accuracy which stood at 92\% gave a good indication that the model learned to classify the unseen reviews well. Also, since the per label F1 Scores for both negative and positive labels stood above 92\%, it gave a good indication that the classification performance on both negative and positive reviews were learned equally well by the model. The evaluation metrics for the test set are reported in the table \ref{table2}.

\begin{table}[ht!]
\caption{The Testing Accuracy and per label Precision, Recall and F-1 Score of the DistilBERT model trained with real labelled sentiment data}
\centering
\begin{tabular}{|l|p{1.3cm}|p{1.3cm}|p{1.3cm}|p{1.3cm}|}
\hline
\textbf{Label}    & \textbf{Per Label   Precision} & \textbf{Per Label Recall} & \textbf{Per Label F-1   Score} & \textbf{Accuracy}                \\ \hline
Negative & 0.9031                & 0.9484           & 0.9252                & \multirow{2}{*}{0.9233} \\ \cline{1-4}
Positive & 0.9457                & 0.8982           & 0.9213                &                         \\ \hline
\end{tabular}
\label{table2}
\end{table}

\begin{table*}[ht!]
\caption{The Intrinsic Evaluation of Generation Models}
\centering
\begin{tabular}{|p{3.5cm}|lll|lll|llll|}
\hline
\multirow{3}{3cm}{Model} &
      \multicolumn{3}{p{3.5cm}|}{\textbf{Metrics for performance with respect to the classifier trained with real data}} &
  \multicolumn{3}{p{3.5cm}|}{\textbf{Metrics for semantic similarity analysis with respect to the inputs used in generation}} &
  \multicolumn{4}{p{3.5cm}|}{\textbf{Metrics to analyse the text quality and diversity of generated data itself}} \\ \cline{2-11} 
 &
  \multicolumn{1}{l|}{\multirow{2}{*}{\textbf{Label}}} &
  \multicolumn{1}{l|}{\multirow{2}{1.5cm}{\textbf{Per Label F-1 Score}}} &
  \multirow{2}{*}{\textbf{Accuracy}} &
  \multicolumn{3}{c|}{\textbf{BERT-Score}} &
  \multicolumn{1}{l|}{\multirow{2}{*}{\textbf{PPL}}} &
  \multicolumn{1}{l|}{\multirow{2}{*}{\textbf{Dist-1}}} &
  \multicolumn{1}{l|}{\multirow{2}{*}{\textbf{Dist-2}}} &
  \multirow{2}{*}{\textbf{Dist-3}} \\ \cline{5-7}
 &
  \multicolumn{1}{l|}{} &
  \multicolumn{1}{l|}{} &
   &
  \multicolumn{1}{l|}{\textbf{Precision}} &
  \multicolumn{1}{l|}{\textbf{Recall}} &
  \textbf{F-1 Score} &
  \multicolumn{1}{l|}{} &
  \multicolumn{1}{l|}{} &
  \multicolumn{1}{l|}{} &
   \\ \hline
\multirow{2}{3cm}{GPT-2 with a single prompt} &
  \multicolumn{1}{l|}{Negative} &
  \multicolumn{1}{l|}{0.7560} &
  \multirow{2}{*}{0.7468} &
  \multicolumn{1}{l|}{\multirow{2}{*}{0.7222}} &
  \multicolumn{1}{l|}{\multirow{2}{*}{0.7291}} &
  \multirow{2}{*}{0.7256} &
  \multicolumn{1}{l|}{\multirow{2}{*}{387.21}} &
  \multicolumn{1}{l|}{\multirow{2}{*}{0.1498}} &
  \multicolumn{1}{l|}{\multirow{2}{*}{0.6598}} &
  \multirow{2}{*}{0.8728} \\ \cline{2-3}
 &
  \multicolumn{1}{l|}{Positive} &
  \multicolumn{1}{l|}{0.7369} &
   &
  \multicolumn{1}{l|}{} &
  \multicolumn{1}{l|}{} &
   &
  \multicolumn{1}{l|}{} &
  \multicolumn{1}{l|}{} &
  \multicolumn{1}{l|}{} &
   \\ \hline
\multirowcell{3}[1ex][l]{T5 with a single prompt at \\ encoder level} &
  \multicolumn{1}{l|}{Negative} &
  \multicolumn{1}{l|}{0.7289} &
  \multirow{2}{*}{0.7690} &
  \multicolumn{1}{l|}{\multirow{2}{*}{0.7216}} &
  \multicolumn{1}{l|}{\multirow{2}{*}{0.7623}} &
  \multirow{2}{*}{0.7411} &
  \multicolumn{1}{l|}{\multirow{2}{*}{1.0851}} &
  \multicolumn{1}{l|}{\multirow{2}{*}{0.0306}} &
  \multicolumn{1}{l|}{\multirow{2}{*}{0.1563}} &
  \multirow{2}{*}{0.3071} \\ \cline{2-3}
 &
  \multicolumn{1}{l|}{Positive} &
  \multicolumn{1}{l|}{0.7988} &
   &
  \multicolumn{1}{l|}{} &
  \multicolumn{1}{l|}{} &
   &
  \multicolumn{1}{l|}{} &
  \multicolumn{1}{l|}{} &
  \multicolumn{1}{l|}{} &
   \\ \Xhline{3\arrayrulewidth}
\multirow{2}{3.5cm}{T5 only with a single prompt \\ at decoder level} &
  \multicolumn{1}{l|}{Negative} &
  \multicolumn{1}{l|}{0.7132} &
  \multirow{2}{*}{0.7447} &
  \multicolumn{1}{l|}{\multirow{2}{*}{0.7600}} &
  \multicolumn{1}{l|}{\multirow{2}{*}{0.7842}} &
  \multirow{2}{*}{0.7716} &
  \multicolumn{1}{l|}{\multirow{2}{*}{1.1533}} &
  \multicolumn{1}{l|}{\multirow{2}{*}{0.0702}} &
  \multicolumn{1}{l|}{\multirow{2}{*}{0.3669}} &
  \multirow{2}{*}{0.6330} \\ \cline{2-3}
 &
  \multicolumn{1}{l|}{Positive} &
  \multicolumn{1}{l|}{0.7700} &
   &
  \multicolumn{1}{l|}{} &
  \multicolumn{1}{l|}{} &
   &
  \multicolumn{1}{l|}{} &
  \multicolumn{1}{l|}{} &
  \multicolumn{1}{l|}{} &
   \\ \hline
\multirow{2}{3.5cm}{T5 with prompts at encoder \\ and decoder levels both} &
  \multicolumn{1}{l|}{Negative} &
  \multicolumn{1}{l|}{0.7284} &
  \multirow{2}{*}{0.7819} &
  \multicolumn{1}{l|}{\multirow{2}{*}{0.7340}} &
  \multicolumn{1}{l|}{\multirow{2}{*}{0.7707}} &
  \multirow{2}{*}{0.7516} &
  \multicolumn{1}{l|}{\multirow{2}{*}{1.0884}} &
  \multicolumn{1}{l|}{\multirow{2}{*}{0.0363}} &
  \multicolumn{1}{l|}{\multirow{2}{*}{0.1632}} &
  \multirow{2}{*}{0.2967} \\ \cline{2-3}
 &
  \multicolumn{1}{l|}{Positive} &
  \multicolumn{1}{l|}{0.8178} &
   &
  \multicolumn{1}{l|}{} &
  \multicolumn{1}{l|}{} &
   &
  \multicolumn{1}{l|}{} &
  \multicolumn{1}{l|}{} &
  \multicolumn{1}{l|}{} &
   \\ \Xhline{3\arrayrulewidth}
\multirowcell{3}[1ex][l]{T5 encoder- decoder prompts\\ with steering at decoder level} &
  \multicolumn{1}{l|}{Negative} &
  \multicolumn{1}{l|}{0.9180} &
  \multirow{2}{*}{0.9185} &
  \multicolumn{1}{l|}{\multirow{2}{*}{0.7345}} &
  \multicolumn{1}{l|}{\multirow{2}{*}{0.7620}} &
  \multirow{2}{*}{0.7472} &
  \multicolumn{1}{l|}{\multirow{2}{*}{1.1725}} &
  \multicolumn{1}{l|}{\multirow{2}{*}{0.1503}} &
  \multicolumn{1}{l|}{\multirow{2}{*}{0.5233}} &
  \multirow{2}{*}{0.7270} \\ \cline{2-3}
 &
  \multicolumn{1}{l|}{Positive} &
  \multicolumn{1}{l|}{0.9190} &
   &
  \multicolumn{1}{l|}{} &
  \multicolumn{1}{l|}{} &
   &
  \multicolumn{1}{l|}{} &
  \multicolumn{1}{l|}{} &
  \multicolumn{1}{l|}{} &
   \\ \hline
\end{tabular}
\label{table3}
\end{table*}

\subsection{Evaluation of Generation Models and Generated Data}

With the experiments related to generation models with soft prompts, our main aim was to preserve the given sentiment of the generated data compared to the input data and also to generate data that are not too different from the input references. For this purpose, we used the GPT-2 with the input soft prompt as discussed by \citep{yang2022tailor} and T5 model with just one soft prompt at the encoder level as discussed by \citep{lester2021power} as our baselines and tried to investigate the usage of soft prompt at decoder level of the T5 model in 2 scenarios: (a) only at the decoder level of T5 model and, (b) at both encoder and decoder levels of the T5 model, as explained in the  model formulation section. Further, the steering strategy at decoder level as discussed in the methodology section was tried upon the T5 model with encoder and decoder level soft  prompts. 

\subsubsection{Intrinsic Evaluation}

In our experiments with regard to intrinsic evaluation, we try to analyse the created generation models with respect to 3 main evaluation criteria: (a) performance of artificial review generation with respect to the sentiment classifier trained with real labelled data, (b) semantic similarity analysis with respect to the inputs used in generation, and (c) the text quality and diversity of generation. 

In order to evaluate the performance of the generation of reviews with respect to the sentiment classifier trained with real labelled data, we use the per label F-1 Scores to see how well the classification happens with respect to each label and the classification accuracy as the overall metric. Even in the baseline, T5 generated data seem to perform better than the GPT-2 generated data when observing the overall accuracy in table \ref{table3} and the generations from T5 model with both encoder and decoder soft prompted model seem to outperform the generations from both of the baseline models (single soft prompted GPT2 and T5) when looking at the overall classification accuracy. This may be due to the fact that the soft prompt at decoder level gives extra help in generation as we hypothesized.  Although the steering at decoder level for this T5 model with both encoder and decoder soft prompts seem to give the best classification accuracy, when eyeballing the generated data, it was observed that the overall quality including the grammatical correctness was quite poor in this setting comparatively. Also, it was observed that long repetitions of articles such as "is is is is" , " was was was was" and nuance token repetitions such as "disc r disc r ", "20th 20th 20th ", and etc were quite frequently prevalent in the generated data from the steering setting.

To analyse the semantic similarity with the input text used for generations, BERT-Score \citep{zhang2019bertscore} was used as this measure gave an overall idea on how close the generated text are compared to the input text used for this generation as we want our generated text to preserve the movie and TV related review context and not be too different from the input text semantically rather than syntactically. As observed from table \ref{table3} highest F-1 Score in BERT-Score is observed in the generations produced by the T5 model with decoder only soft prompt followed by the generations of the T5 model with both encoder and decoder soft prompts which are above 75\% .

When it comes to the text quality and diversity analysis of the generations from models, the lowest perplexity (PPL) in table \ref{table3} is observed with the generations from T5 model with only encoder soft prompt followed by T5 model with both encoder and decoder soft prompts only with a difference of 0.0033 in the score, which depicts that our new model is almost equally confident and fluent in text generalization compared with the T5 model with an only encoder prompt. The diversity of the generations were measured following the method discussed by \citep{li2016diversity} where the number of unigrams, bigrams and trigrams are counted and normalized (i.e., Dist-1 / Dist-2 / Dist-3) to check if the generations are distinct enough not to generate the repetitions of token strings but not too diverse as well to swing away from the context of generation. In this case we notice that our T5 model with both encoder and decoder soft prompts gives quite a low score compared to the generations from other models.

\subsubsection{Memorisation Evaluation}

Since the soft prompts are finetuned to preserve the context of training data, there is an actual risk of the models just repeating the training data phrases during generation. To check this, unique n-grams from training set of the generation models were compared against the unique n-grams of the generated text, to find the overlap between these 2 sets. In general when the "n" of the n-grams increases the overlap percentage should decrease if the models are not just repeating training data as the probability of having common n-grams should decrease when the length of the n-grams increase.

\begin{table}[ht!]
\caption{n-gram overlap analysis between the generated data and the training data used for generation models}
\centering
\begin{tabular}{|p{2cm}|p{1cm}|p{1cm}|p{1cm}|p{1cm}|}
\hline
\textbf{Model}    & \textbf{2-gram overlap (\%)} & \textbf{3-gram overlap (\%)} & \textbf{4-gram overlap (\%)} & \textbf{5-gram overlap (\%)} \\ \hline
Prompted GPT-2                              & \makecell*[c{c}]{26.89} & \makecell*[c{c}]{11.66} & \makecell*[c{c}]{3.21}  & \makecell*[c{c}]{0.67} \\ \hline
T5 only with encoder prompt            & \makecell*[c{c}]{43.24} & \makecell*[c{c}]{21.97} & \makecell*[c{c}]{8.57}  & \makecell*[c{c}]{2.76} \\ \hline
T5 only with   decoder prompt      & \makecell*[c{c}]{34.91} & \makecell*[c{c}]{17.38} & \makecell*[c{c}]{6.63}  & \makecell*[c{c}]{2.04} \\ \hline
T5 with  encoder - decoder prompts & \makecell*[c{c}]{45.85} & \makecell*[c{c}]{24.71} & \makecell*[c{c}]{10.05} & \makecell*[c{c}]{3.40} \\ \hline
T5 encoder - decoder prompts with steering at decoder level & \makecell*[c{c}]{25.67}               & \makecell*[c{c}]{11.20} & \makecell*[c{c}]{3.62}  & \makecell*[c{c}]{0.99}                \\ \hline
\end{tabular}
\label{table4}
\end{table}

The results are reported in the table \ref{table4} and although according to table \ref{table4} the overlap percentages are decreasing when the lengths of the n-grams are increasing, it is observed that the unique n-gram overlap percentages for the generations from our T5 model with both encoder and decoder soft prompts is higher in general compared to the generations from other models. Therefore to see if important information, specially nouns are leaking from training data to generated data, the common overlaps of the n-grams of training and generated data for this model were eyeballed and it was observed that mostly the overlaps are article-adverb-verb phrases like "definitely\_watch" "did\_see", \& "wasted\_on" for 2-grams, "to\_be\_awesome", "are\_not\_likable" \& "the\_kind\_of" for 3-grams, "has\_so\_much\_to" \& "but\_it\_was\_the" for 4-grams and "it\_is\_a\_waste\_of" \& "this\_would\_have\_been\_a" for 5-grams etc., but not any proper-nouns related phrases. The results of the unique n-gram overlap analysis between the generated data and the testing data used in generation (reported in Appendix \ref{ap2} Section \ref{ap21}) also gave similar observations.

\subsubsection{Extrinsic Evaluation}

Extrinsic evaluation of the generated data was performed by finetuning 3 separate DistilBERT models with generated data from the three T5 based models with soft prompts which performed well in the intrinsic evaluation and those finetuned classifier models were tested on the same real labelled data test set from Dataset-1 which was used to evaluate the sentiment classifier fine tuned with real labelled data.

\begin{table}[ht!]
\caption{The Testing Accuracy, Precision, Recall and F-1 Score of the DistilBERT models trained with generated sentiment data}
\centering
\begin{tabular}{|p{1.8cm}|p{0.8cm}|p{0.9cm}|p{0.9cm}|p{0.9cm}|p{0.9cm}|}
\hline
\multirowcell{2}[-2ex][c]{\textbf{Model}} & \multirowcell{2}[-2ex][c]{\textbf{Label}} & \multicolumn{3}{c|}{\textbf{Per Label}}                                             & \multirowcell{2}[-2ex][l]{\textbf{Accuracy}} \\ \cline{3-5}
                       &                        & \multicolumn{1}{l|}{\multirowcell{2}[0pt][c]{\textbf{Precision}}} & \multicolumn{1}{l|}{\multirowcell{2}[0pt][c]{\textbf{Recall}}} & \textbf{F-1   Score} &                           \\ \hline
\multirow{2}{2cm}{Trained with generations from T5 model with an encoder prompt only} &
  \multirow{3}{*}{Negative} &
  \multirow{3}{*}{0.7839} &
  \multirow{3}{*}{0.9332} &
  \multirow{3}{*}{0.8521} &
  \multirow{6}{*}{0.8380} \\ &&&&& \\ &&&&& \\ \cline{2-5}
 &
  \multirow{2}{*}{Positive} &
  \multirow{2}{*}{0.9175} &
  \multirow{2}{*}{0.7428} &
  \multirow{2}{*}{0.8210} &
   \\ &&&&& \\  \hline
\multirow{2}{2cm}{Trained with generations from T5  model with encoder \& decoder prompts} &
  \multirow{3}{*}{Negative} &
  \multirow{3}{*}{0.8507} &
  \multirow{3}{*}{0.9114} &
  \multirow{3}{*}{\textbf{0.8800}} &
  \multirow{5}{*}{\textbf{0.8757}} \\ &&&&& \\ &&&&& \\ \cline{2-5}
 &
  \multirow{2}{*}{Positive} &
  \multirow{2}{*}{0.9046} &
  \multirow{2}{*}{0.8400} &
  \multirow{2}{*}{\textbf{0.8711}} &
   \\ &&&&& \\ \hline
\multirow{2}{2cm}{Trained with generations from T5  model with encoder \& decoder prompts with steering} &
  \multirow{3}{*}{Negative} &
  \multirow{3}{*}{0.7924} &
  \multirow{3}{*}{0.9568} &
  \multirow{3}{*}{0.8669} &
  \multirow{6}{*}{0.8531} \\ &&&&& \\ &&&&& \\ \cline{2-5}
 &
  \multirow{3}{*}{Positive} &
  \multirow{3}{*}{0.9455} &
  \multirow{3}{*}{0.7494} &
  \multirow{3}{*}{0.8361} &
   \\ &&&&& \\ &&&&& \\ \hline
\end{tabular}
\label{table5}
\end{table}

The results from the extrinsic evaluation which are reported in the table \ref{table5} showed that the data generated from our T5 model with both encoder and decoder soft prompts gave the best classification evaluation results compared to all the generations from other models, with a 87.57\% accuracy and around 87\% per label F-1 Scores for both negative and positive labelled data, with only a decrease of 4.7\% in accuracy compared to the classifier trained with real labelled data. Data generated from steering the T5 model with both encoder and decoder soft prompts seem to under-perform compared to generations from our main proposed model, most probably due to the grammatical and structural errors in the generated data from this model identified during the manual analysis performed in intrinsic evaluation. 

This extrinsic evaluation gives a view that the generations from our proposed T5 model with both encoder and decoder soft prompts can be utilized for AI tasks such as training classifier models to compensate the lack of relevant real labelled data without compensating very much on the levels of classification accuracy. 
.

\subsection{Interpretability Evaluation of the Classifier Model trained with Artificially Generated Data}

\begin{figure}[htbp]
\centerline{\includegraphics[height=0.48\linewidth]{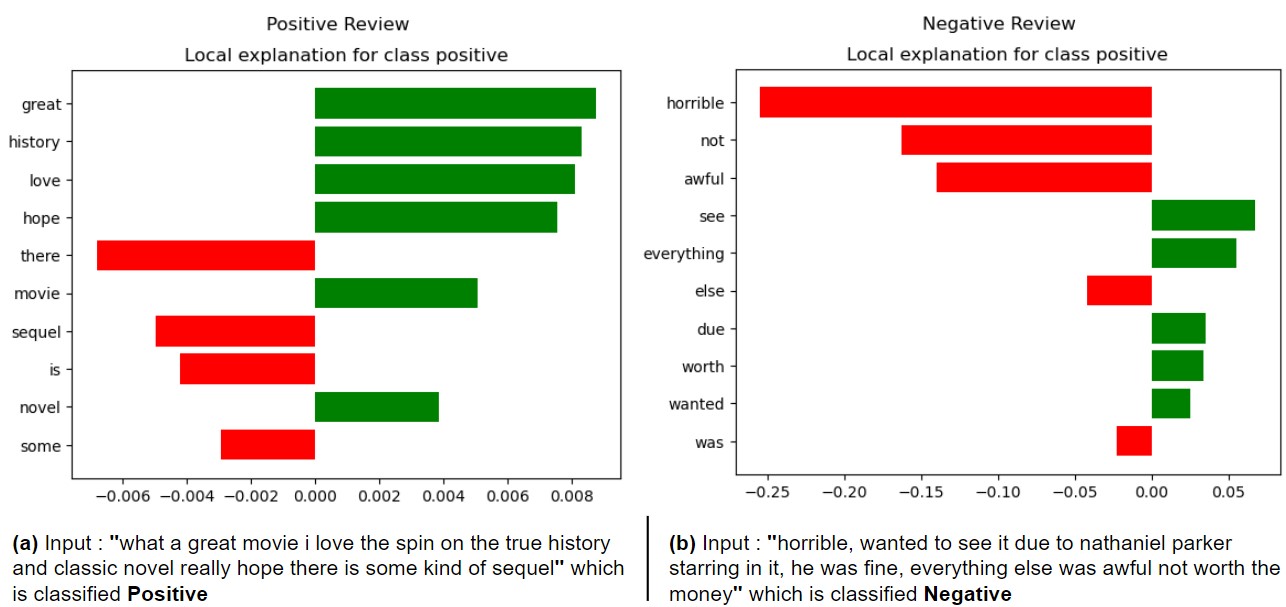}}
\caption{The interpretability analysis of the classifier model trained with generated data using the LIME model}
\label{fig4}
\end{figure}

In order to investigate further what features of the input data drives the classification decision by the classification model trained with artificially generated data from our T5 model with both encoder and decoder soft prompts, we performed an analysis using the LIME technique \citep{ribeiro2016should} which is a model agnostic methodology which perturb the input text around its neighborhood to try to check how the classifier model's predictions behave and learn an interpretable new model based on those perturbations on top of our finetuned classifier model, to give a score to each word in the input on how much they contributed for the classification decision.   

For the tested examples reported in Figure \ref{fig4}, we can observe that for the positive classification in part (a), the words such as "great", "love", and "hope" has driven the classification decision which are considered positive in general by humans as well, while for the negative classification in part (b), the words such as "horrible", "not", and "awful" has contributed mostly for the classification decision, which are considered negative in general according to the LIME model learned on top of our finetuned classifier.

This analysis shows that the classifier trained with artificial data makes the classification decision based on interpretable features of the input where in this case can be identified as positive and negative words of the input text in general.

\section{Conclusions and Future Work}

In this work, we mainly introduce a novel extended soft prompt tuning architecture for T5 model incorporating soft prompts at both encoder and decoder levels, to produce controlled text generation and investigate its performance. Through the analysis of the results obtained by the intrinsic and specially by the extrinsic evaluation of the data produced by this extended model, it seems that this extended model produced promising and superior results compared to the results produced by the base line models, for the task of generating sentiment preserved TV and movie related reviews. But it is also observed that it does so, with the expense of more memorization (of mainly grammatical structures) compared to the other models.

Even though steering at decoder level of this newly formulated encoder-decoder soft prompted T5 model produced better results in the intrinsic evaluation compared to the main proposed model, it seems steering of an encoder-decoder soft prompted model needs more investigations as the grammatical quality of the generated text was found to be sub-par compared to other generations and also due to the fact that it couldn't outperform the results of the main proposed model in extrinsic evaluation. 

Also, the results obtained via the interpretability evaluation of the classifier trained with generated data along with the outcomes of the extrinsic evaluation of the generated data, suggests that these generated texts may be used in AI tasks such as model training, as this model trained with generated data produced classification results comparable to the results of a classifier trained with real labelled data where this classifier trained with artificial data also seemed to make the classification decisions based on features of the inputs which seemed to be interpretable in general context.

For future work, an in-depth analysis of the extent of memorization taking place in our proposed main model could be performed along with a further research study of methods and techniques which can be incorporated to minimize this effect maybe in a privacy preserving perspective.

\section*{Acknowledgment}

We'd like to express our heartiest gratitude to the School of Electronic Engineering and Computer Science of Queen Mary University of London for providing us the access to infrastructure such as the access to the GPU cluster to conduct our experiments.


\bibliographystyle{agsm}
\bibliography{bibfile}

\appendices

\section{Model Formulation Details}
\label{ap0}

\subsection{Expert \& Anti Expert Steering details for the T5 model with Encoder-Decoder Soft Prompts}
\label{ap01}

\begin{figure}[htbp]
\centerline{\includegraphics[height=0.8\linewidth]{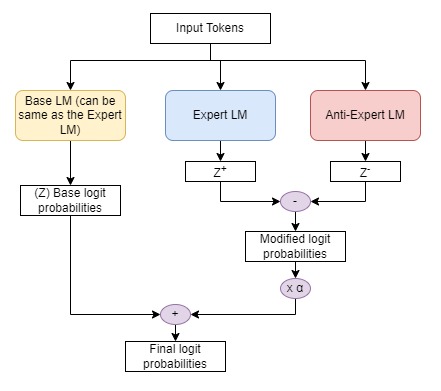}}
\caption{The Block Diagram for the Working Principle of Steering the output of T5 model with encoder and decoder soft prompts}
\label{fig3}
\end{figure}

\section{Experiment Details}
\label{ap1}

\subsection{Platform Information}
\label{ap11}

All the experiments were conducted in the QMUL EECS JHub platform utilizing the Nvidia A100 GPUs and the RAM available with the platform for use was 12GB. The version of the Python used was 3.10.6.

\subsection{Sentiment Classifier Model Training with Real Data}
\label{ap12}

This DistilBERT model was finetuned with the Dataset-1 which was prepared with a training set of 50K examples (25K positive examples and 25K negative examples) and validated and tested with the Dataset-1’s 10K (5K negative and 5K positive) validation and testing splits. The hyperparmeters reported in the table \ref{table6} were used for training keeping all the other hyperparemeters with their default values as set by the Hugging Face library and the PyTorch framework.

\begin{table}[ht!]
\centering
\caption{Hyperparameters for the classification model training}
\label{table6}
\begin{tabular}{l|l}
\textbf{Hyperparameter} & \textbf{Values}   \\
\hline
learning rate  & 1e-05 \\
batch size     & 32       \\
no. of epochs  & 5        \\
seed           & 123    \\
\end{tabular}
\end{table}

The model which gave the best validation accuracy (it was at epoch 3) was chosen to produce test metrics and this same model was used for all the data testing purposes in the experiments afterwards.

\subsection{Generation Model Training}
\label{ap13}

A soft prompt token with a token length of 20 tokens was prepended at the input layer of the GPT-2 model and was trained with the Dataset-2's training set and then validated and tested on Dataset-2's validation and testing sets respectively. 

Then the T5 models with encoder only prompt, decoder only prompt and encoder and decoder prompts were also trained similarly where the soft prompt token length was fixed at 20 tokens with the Dataset-2's training set and then validated and tested on Dataset-2's validation and testing sets respectively. 

\begin{table}[ht!]
\centering
\caption{Hyperparameters related to Adafactor Optimizer used for the training of Generation Models}
\label{table7}
\begin{tabular}{l|l}
\textbf{Hyperparameter} & \textbf{Values}   \\
\hline
eps  & (1e-30, 1e-3) \\
clip\_threshold     & 1.0       \\
decay\_rate  & -0.8        \\
beta1 & None    \\
weight\_decay & 0.1 \\
relative\_step & False \\
scale\_parameter & False \\
warmup\_init & False \\
no. of epochs  & 20 \\
warmup\_steps & 500 \\
batch size     & 10       \\
seed           & 123    \\

\end{tabular}
\end{table}

Adafactor optimizer was chosen for tuning all the soft prompted generation models as per the discussion in \citep{lester2021power} and a learning rate of 5e-5 was chosen for the GPT-2 soft prompted model according to the discussion in \citep{yang2022tailor} and a learning rate of 0.15 was chosen for all the T5 related generation models according to the insights gained from the work of \citep{lester2021power}. The common hyperparameters related to the Adafactor optimizer used to train these generation models are reported in the table \ref{table7}. The batch size was selected as 10 due to the restrictions in GPU memory availability during training.

The model which gave the lowest validation loss was chosen to produce test metrics and this same model was used for all the data generation purposes in the experiments afterwards.

\begin{table}[ht!]
\centering
\caption{Parameters related to text generation through the trained models at inference stage using Top-p (nucleus) sampling}
\label{table8}
\begin{tabular}{l|l}
\textbf{Parameter} & \textbf{Values}   \\
\hline
num\_beams & 10 \\ 
do\_sample & True \\
no\_repeat\_ngram\_size & 1 \\  
temperature & 1.0 \\
top\_k & 0 \\
top\_p & 0.8 \\
repetition\_penalty & 1.0 \\
use\_cache & False \\
early\_stopping & True \\

\end{tabular}
\end{table}

As the decoding strategy in text generation through the obtained models, Top-p (nucleus) sampling methodology was chosen as this method was considered the best according to the discussion in \citep{holtzman2019curious} for large language models. The parameters used in this decoding strategy during text generation are reported in the table \ref{table8}.

\subsection{Steering Model Setup}
\label{ap14}

The T5 encoder-decoder soft prompted models with lowest validation loss for positive and negative review generations were chosen, and for positive review generation the T5 encoder-decoder soft prompted model trained for positive review generation was used as the base and the expert and the T5 encoder-decoder soft prompted model trained for negative review generation was used as the anti expert as shown in the figure \ref{fig3}. For the negative review generation the T5 encoder-decoder soft prompted model trained for negative review generation was used as the base and the expert model while T5 encoder-decoder soft prompted model trained for positive review generation was used as the anti-expert. 

1.2 was used for $\alpha$ which is the hyperparameter of this steering model as seen in the equation \ref{eq1}

The same Top-p (nucleus) sampling methodology was used as the decoding strategy in text generation using this decoder level steered model as well and the parameters used for the positive and negative text generations are reported in the table \ref{table9}

\begin{table}[ht!]
\centering
\caption{Parameters related to text generation through the T5 steering at inference stage using Top-p (nucleus) sampling}
\label{table9}
\begin{tabular}{l|p{2cm}|p{2cm}}
\textbf{Parameter} & \textbf{Values for Positive Steering} & \textbf{Values for Negative steering}  \\
\hline
sample & True & True \\
filter\_p  &  1 & 0.9 \\
k & 0 & 0 \\
p & 0.9 & 0.9  \\
temperature & 1.1 & 1.8 \\
alpha & 1.2 & 1.2 \\

\end{tabular}
\end{table}

\subsection{Classifier Model Training with Artificial Data}

Separate DistilBERT models were finetuned with the generated data from different generation models where separate training sets of 50K examples (25K positive examples and 25K negative examples) were generated using those generation models taking the training split of Dataset-3 as the input and were validated with the generated 10K (5K negative and 5K positive) validation data generated using the validation split of Dataset-3 as the input. Testing for classification was done with the testing split of the Dataset-1  to compare the classification results among these models and the classifier trained with real labelled data. The same hyperparmeters reported in the table \ref{table6} were used for training, keeping all the other hyperparemeters with their default values as set by the Hugging Face library and the PyTorch framework.

Similar to the strategy  in  the Appendix \ref{ap1} Section \ref{ap12}, the models which gave the best validation accuracy were chosen to produce test metrics.

\section{Evaluation Analysis Details}
\label{ap2}

\subsection{Unique N-gram Overlap Analysis Details between the Generated Data and the Testing Data used for Generation}
\label{ap21}

\begin{table}[ht!]
\caption{Unique n-gram overlap analysis details between the generated data and the testing data used for generation}
\centering
\begin{tabular}{|p{2cm}|p{1cm}|p{1cm}|p{1cm}|p{1cm}|}
\hline
\textbf{Model}    & \textbf{2-gram overlap (\%)} & \textbf{3-gram overlap (\%)} & \textbf{4-gram overlap (\%)} & \textbf{5-gram overlap (\%)} \\ \hline
Prompted GPT-2                              & \makecell*[c{c}]{14.87} & \makecell*[c{c}]{5.71} & \makecell*[c{c}]{1.38}  & \makecell*[c{c}]{0.25} \\ \hline
T5 only with encoder  prompt            & \makecell*[c{c}]{41.73} & \makecell*[c{c}]{24.64} & \makecell*[c{c}]{14.17}  & \makecell*[c{c}]{8.74} \\ \hline
T5 only with   decoder prompt      & \makecell*[c{c}]{28.54} & \makecell*[c{c}]{16.97} & \makecell*[c{c}]{10.14}  & \makecell*[c{c}]{6.53} \\ \hline
T5 with   encoder - decoder prompts & \makecell*[c{c}]{48.14} & \makecell*[c{c}]{31.26} & \makecell*[c{c}]{19.13} & \makecell*[c{c}]{12.26} \\ \hline
T5 encoder - decoder prompts with steering at decoder level & \makecell*[c{c}]{18.83}          & \makecell*[c{c}]{12.11}  & \makecell*[c{c}]{8.60}  &  \makecell*[c{c}]{6.72}                \\ \hline
\end{tabular}
\label{table10}
\end{table}

\subsection{Overlapped N-gram Examples for Generated Data vs Model Training Data and Other Evaluation Analysis Details}
\label{ap22}

\begin{table}[ht!]
\centering
\caption{Overlapped 2-gram and 3-gram examples for the generated data from T5 with encoder and decoder soft prompts and its training data}
\label{table11}
\begin{tabular}{p{3.5cm}|p{3.5cm}}
\textbf{2-gram examples} & \textbf{3-grams examples}   \\
\hline
week\_but, dialog\_was, definitely\_watch, moments\_throughout, job\_in, am\_sure, is\_true, does\_end, want\_these, asleep\_in    &     the\_point\_of, get\_through\_it, besides\_that\_the, to\_be\_awesome, think\_that\_would, was\_not\_clear, of\_all\_movies, what\_you\_are, sure\_you\_can, check\_it\_out     \\

\end{tabular}
\end{table}

\begin{table}[ht!]
\centering
\caption{Overlapped 3-gram and 4-gram examples for the generated data from T5 with encoder and decoder soft prompts  and its training data}
\label{table12}
\begin{tabular}{p{3.5cm}|p{3.5cm}}
 \textbf{4-grams examples} & \textbf{5-grams examples}  \\
\hline
movies\_that\_could\_have, has\_so\_much\_to, but\_it\_was\_the, as\_it\_is\_not, a\_remake\_of\_the, interesting\_twist\_on\_the, we\_stopped\_watching\_it, waste\_my\_time\_and, to\_believe\_that\_the    &     this\_would\_have\_been\_a, it\_is\_a\_waste\_of, this\_would\_be\_a\_good, this\_is\_a\_great\_example, recommend\_this\_for\_anyone\_who, had\_a\_great\_cast\_and, but\_it\_did\_not\_make, love\_this\_movie\_as\_a, i\_think\_it\_was\_a, plot\_was\_just\_a\_little    \\

\end{tabular}
\end{table}

\begin{figure*}[htbp]
\centerline{\includegraphics[height=0.3\linewidth]{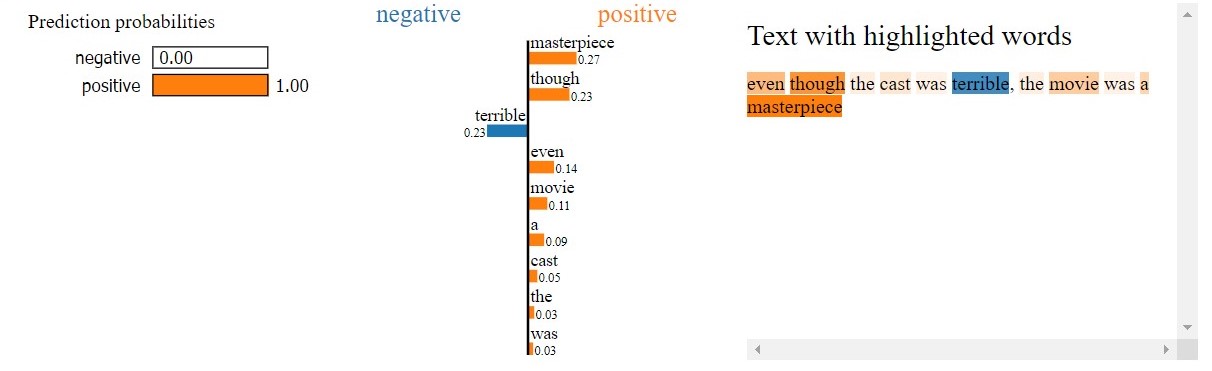}}
\caption{The interpretability analysis of the classifier model trained with generated data using the LIME model for an input text classified Positive}
\label{fig5}
\end{figure*}

\begin{figure*}[htbp]
\centerline{\includegraphics[height=0.3\linewidth]{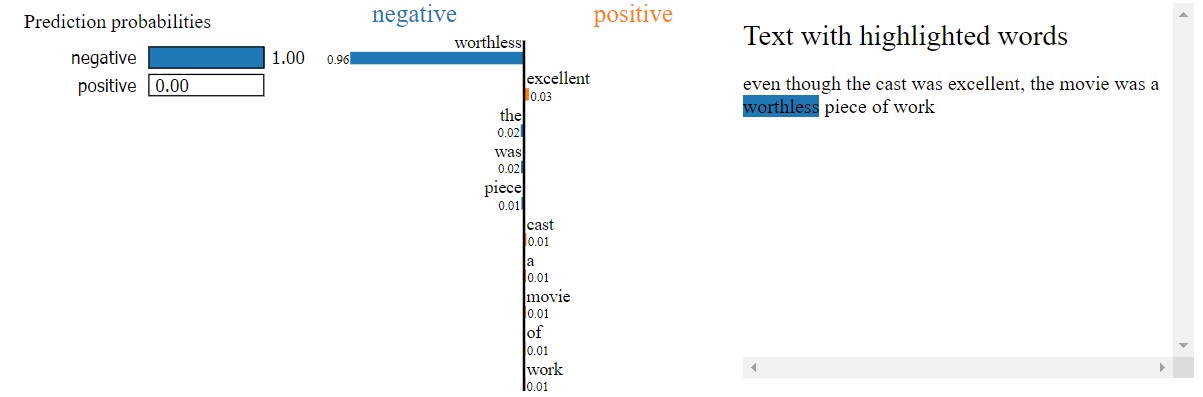}}
\caption{The interpretability analysis of the classifier model trained with generated data using the LIME model for an input text classified Negative}
\label{fig6}
\end{figure*}

\vspace{12pt}

\end{document}